%% file: samplepaper.tex
\documentclass[runningheads]{llncs}
\usepackage[T1]{fontenc}
\usepackage{amsmath}
\usepackage{amsfonts}
\usepackage{array}
\usepackage{cellspace}
\usepackage[multiple]{footmisc}
\usepackage{algorithm}
\usepackage{algorithmicx}
\usepackage{algpseudocode}
\usepackage{listings}
\usepackage{graphicx}
\usepackage{orcidlink}
\begin{document}
\title{U-CESE: Unified Clip-based Event Search Engine \\for AI Challenge HCMC 2025}
\titlerunning{U-CESE: Unified Clip-based Event Search Engine}
%

\newcommand{\footnoteEqCon}{\protect\footnote[2]{These authors contributed equally to this research.}}
\newcommand{\footnotemarkEqCon}{\protect\footnotemark[2]{}}
\newcommand{\footnoteCorAuth}{\protect\footnote[1]{Corresponding author(s). E-mail(s): \email{ldnhuan22@apcs.fitus.edu.vn}}}

\author{
Duc-Nhuan Le\footnoteCorAuth\footnoteEqCon\inst{1,2}\orcidlink{0009-0007-9611-0114} \and
Hoang-Phuc Nguyen\footnotemarkEqCon\inst{1,2}\orcidlink{0009-0000-2946-7956} \and
Thanh-Duy Lam\footnotemarkEqCon\inst{1,2}\orcidlink{0009-0009-4443-4082} \and \\
Minh-Nhut Dang\footnotemarkEqCon\inst{1,2}\orcidlink{0009-0007-7162-1727} \and
Minh-Hoang Le\footnotemarkEqCon\inst{1,2}\orcidlink{0009-0005-1501-8080}
}
\authorrunning{Nhuan et al.}
%
\institute{Faculty of Information Technology, University of Science, VNU-HCM \and
Vietnam National University, Ho Chi Minh City, Vietnam
\email{\{ldnhuan22,nhphuc222,ltduy22,dmnhut22,lmhoang22\}@apcs.fitus.edu.vn}
}
\maketitle              
%
\begin{abstract}
Retrieving events from large-scale video datasets is challenging due to complex temporal, spatial, and multimodal information. This paper presents U-CESE, our solution for the AI Challenge HCMC 2025, a Unified Clip-based Event Search Engine for multimodal event retrieval across diverse video sources. Building on CESE, U-CESE integrates its three modules into a single cohesive framework, ensuring consistent processing and retrieval across query types. A core component is the Unified Clipping Algorithm, which merges separate clipping algorithms into one efficient pipeline. To handle large-scale data, we propose DAKE, a lightweight, training-free keyframe extraction method using JPEG file size variations to identify significant scene changes. Finally, we introduce ReCap, a temporally consistent captioning framework inspired by Recurrent Neural Network, generating detailed and context-aware textual descriptions. Experiments show that U-CESE delivers robust, consistent, and efficient performance in large-scale multimodal event retrieval.

\keywords{Video Event Retrieval, Text-to-Image Matching, Visual Question Answering}
\end{abstract}

\input{Introduction/Introduction}
\input{RelatedWork/RelatedWork}

\input{ChallengeOverview/ChallengeOverview}
\input{SystemOverview/SystemOverview}

\input{AblationStudy/AblationStudy}

\input{Conclusion/Conclusion}

\section*{Acknowledgments}
This research is supported by research funding from Faculty of Information Technology,
University of Science, Vietnam National University - Ho Chi Minh City.
\bibliographystyle{splncs04}
\bibliography{references}
\end{document}

%% file: Introduction/Introduction.tex
\section{Introduction}

\begin{figure}[h]
    \centering
    \includegraphics[width=1\linewidth]{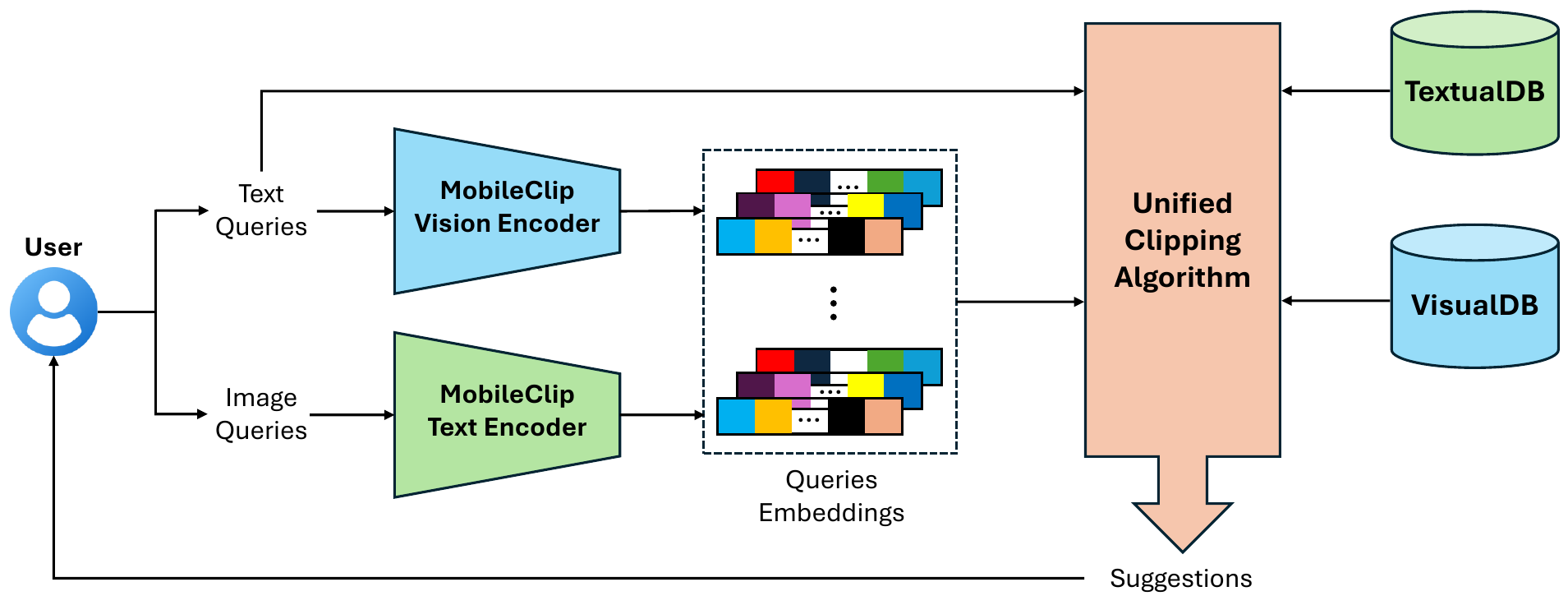}
    \caption{Overall system architecture of U-CESE}
    \label{fig:U-CESE Pipeline}
\end{figure}

The AI Challenge HCMC (AIC) 2025 \cite{AIChallenge2025} tasks participants with developing systems for retrieving specific events from diverse video sources. Building on the structure of the Video Browser Showdown (VBS)~\cite{lokovc2021reign} and the Lifelog Search Challenge (LSC)~\cite{gurrin2023introduction}, AIC 2025 \cite{AIChallenge2025} retains the previous format while introducing a new query type requiring the retrieval of multiple scenes within a video. The dataset has also been significantly expanded in scale and diversity, demanding efficient extraction and utilization of spatial, temporal, and object-level features from large-scale video collections. These capabilities are particularly valuable for navigating extensive TV news archives, where rapid access to relevant visual content enhances search and analysis workflows~\cite{do2023news}.

To address these challenges, we present \textbf{U-CESE}, a \textit{Unified Clip-based Event Search Engine} for AIC  \cite{AIChallenge2025}. Our system extends the CESE framework~\cite{cese}, which retrieves coherent clips matching event descriptions across multiple queries rather than single frames. However, CESE employs three separate modules, each with distinct user interfaces and re-ranking strategies, leading to inconsistency and inefficiency. U-CESE overcomes these limitations by unifying all modalities into a single cohesive module. As illustrated in Fig.~\ref{fig:U-CESE Pipeline}, user queries are processed through our \textbf{Unified Clipping Algorithm}, which replaces CESE’s multiple clipping procedures with a single streamlined framework. To process large-scale video data, we further introduce two novel frameworks: \textbf{DAKE}, a training-free keyframe extraction method based on JPEG file sizes that avoids complex shot boundary models~\cite{TransNetv2,zhuautoshot}; and \textbf{ReCap}, a temporally consistent captioning approach inspired by recurrent neural networks (RNN)~\cite{rnn}, enabling context-aware captions for keyframes.

Our main contributions are summarized as follows:
\begin{itemize}
    \item We propose \textbf{U-CESE}, a unified clip-based event search engine that consolidates all modalities into a single retrieval framework.
    \item We design the \textbf{Unified Clipping Algorithm}, merging CESE’s three separate clipping methods into one consistent and efficient pipeline.
    \item We introduce \textbf{DAKE}, a lightweight, training-free keyframe extraction strategy leveraging JPEG file size variance.
    \item We develop \textbf{ReCap}, a temporally consistent captioning model inspired by RNNs for detailed and context-preserving descriptions.
\end{itemize}

%% file: RelatedWork/RelatedWork.tex
\section{Related Work}

In recent years, video retrieval has evolved into a deep learning task integrating visual feature extraction, linguistic feature extraction, and embedding-based retrieval~\cite{zhu2023deep}. For visual feature extraction, FFmpeg\footnote{https://ffmpeg.org} is commonly used for keyframe extraction~\cite{phat2024revimm}, while OpenCLIP~\cite{openclip} serves as a backbone for frame embedding~\cite{phat2024revimm,gia2024addressing,dinh2024transforming}. Additional models include TransNetV2~\cite{TransNetv2} and AutoShot~\cite{zhuautoshot} for shot boundary detection~\cite{dinh2024transforming}, Grounding DINO for object detection~\cite{dinh2024transforming}, and multimodal models such as InternVL~\cite{internvl}, BEiT-3~\cite{beit3}, and BLIP-2~\cite{blip2} for enhanced representation~\cite{gia2024addressing,dinh2024transforming}. For linguistic feature extraction, Whisper~\cite{Whisper} is the primary ASR model used in several systems~\cite{phat2024revimm,gia2024addressing,dinh2024transforming}, while Vintern-1B~\cite{doan2024vintern} is employed for image captioning~\cite{phat2024revimm}. PARSeq~\cite{parseq} is fine-tuned for Vietnamese OCR~\cite{gia2024addressing}, addressing the limitations of DeepSolo~\cite{deepsolo}. For embedding-based retrieval, similarity search is implemented using Elasticsearch~\cite{elasticsearch,phat2024revimm,cese,gia2024addressing}, FAISS~\cite{faiss,phat2024revimm}, Milvus~\cite{2021milvus,gia2024addressing,dinh2024transforming}, and Polars\footnote{https://pola.rs/}~\cite{dinh2024transforming}. Query reformulation with LLMs further enhances retrieval precision~\cite{gia2024addressing}.

CESE~\cite{cese} introduces a clip-based event search approach that focuses on identifying relevant clips rather than individual frames. Queries can contain multiple scene descriptions, and a clip is considered more relevant if it contains more unique scenes. CESE uses three clipping algorithms: Visual Semantic, Textual Semantic, and Keywords Clipping. Each utilizes visual, textual, or keyword cues to generate candidate clips. While this multi-modal design enhances retrieval robustness, CESE suffers from inconsistency and inefficiency as it relies on three separate modules, each with distinct user interfaces and re-ranking strategies, leading to fragmented workflows and limited system scalability.

%% file: ChallengeOverview/ChallengeOverview.tex
\section{Challenge Overview}

\subsection{Dataset Overview}

The dataset in the AI Challenge HCMC 2025 - Event Retrieval from Visual Data consists of 1478 videos from various TV programs with a total duration of 324 hours. For each video in the dataset, the organizer provides its extracting keyframes and metadata. For each keyframe, objects detected from the Faster RCNN pretrained OpenImagesV4 model \cite{faster_rcnn_inception_resnet_v2} and CLIP features from CLIP ViT-B/32 model \cite{radford2021learning} are also provided.


\begin{table}
    \centering
    \caption{The dataset contains 9 TV Programs. The table informs the TV program's name, broadcast period, number of videos, and average duration per video. There are a total of 1478 videos with a total duration of 324 hours.}
    \begin{tabular}{|c|c|c|c|}
         \hline
        \textbf{TV Program} & \textbf{Period} & \textbf{Videos} & \textbf{Mins/video} \\
         \hline
        "HTV 60 Giay" & 08/24 - 08/25 & 665 & 20 \\
         \hline
        National Bicycle Race HCMC TV Cup 2024 & 04/24 & 25 & 6.5 \\
         \hline
        Lion Dance - Cho Lon Cup & 01/24 & 43 & 9 \\
         \hline
        \begin{tabular}{c}Thanh Nien Newspaper:\\"Bi quyet on thi THPT 2024"\end{tabular} & 05/24 - 06/24 & 88 & 25 \\
         \hline
        "Mon ngon moi ngay" & 04/20 - 09/24 & 498 & 5 \\
         \hline
        "Viet Nam di la ghien" Season 3 & 06/24 - 09/24 & 16 & 6 \\
         \hline
        "Tan man Me Kong" & 05/24 - 09/24 & 24 & 19 \\
         \hline
        "Đoi mat Me Kong" & 05/24 - 09/24 & 23 & 18 \\
         \hline
        \begin{tabular}{c}Tuoi Tre Newspaper:\\"Lan toa nang luong tich cuc 2024"\end{tabular} & 08/24 - 09/24 & 96 & 3.5 \\
        \hline
        \textbf{Total (324 hours)} & & \textbf{1478} & 13.4 \\
        \hline
    \end{tabular}
    \vspace{.8em}
    \label{tab:video_data_information}
\end{table}

\subsection{Query types}

The challenge consists of 4 distinct query types:
\begin{itemize}
    \item \textbf{Known-Item Search (KIS):} Teams are required to locate any frame corresponding to a specific moment in the dataset within 5 minutes. Five textual hints are provided progressively at 1-minute intervals.
    \item \textbf{Video Known-Item Search (VKIS):} Instead of textual hints, teams are shown a 20-second video segment from the dataset and must retrieve the same moment within 4 minutes. Recording the video is not allowed.
    \item \textbf{Visual Question Answering (VQA):} Similar to the KIS task, but includes an additional question about the identified moment.
    \item \textbf{Temporal Alignment (TRAKE):} Participants must determine the exact frames corresponding to several key moments of a single action and arrange them in the correct order.
\end{itemize}

%% file: SystemOverview/SystemOverview.tex

\section{System Overview}


This section presents \textbf{U-CESE}, our unified clip-based event search system, illustrated in Fig.~\ref{fig:U-CESE Pipeline}. The pipeline begins with database construction from raw video data, as described in Sec.~\ref{sec:Data Preprocessing}. This process integrates our two novel frameworks: \textbf{DAKE}, which performs dynamic-aware keyframe extraction, and \textbf{ReCap}, which generates temporally consistent shot captions. U-CESE supports both image and text queries. These inputs are embedded using MobileCLIP encoders~\cite{MobileClip}, producing multimodal representations. The embeddings, along with the raw text queries, are then processed by the \textbf{Unified Clipping Algorithm} (Sec.~\ref{sec:unified_clipping}), which retrieves relevant results from the multimodal database. Finally, the retrieved items are re-ranked and displayed through a unified web interface (Sec.~\ref{sec:user-interface}).

\input{SystemOverview/DataPreprocessing/DataPreprocessing}
\input{SystemOverview/UnifiedClippingAlgorithm/UnifiedClippingAlgorithm}
\input{SystemOverview/UserInterface/UserInterface}

%% file: SystemOverview/DataPreprocessing/DataPreprocessing.tex
\subsection{Data Preprocessing Pipeline}
\label{sec:Data Preprocessing}

\begin{figure}
    \centering
    \includegraphics[width=1.0\linewidth]{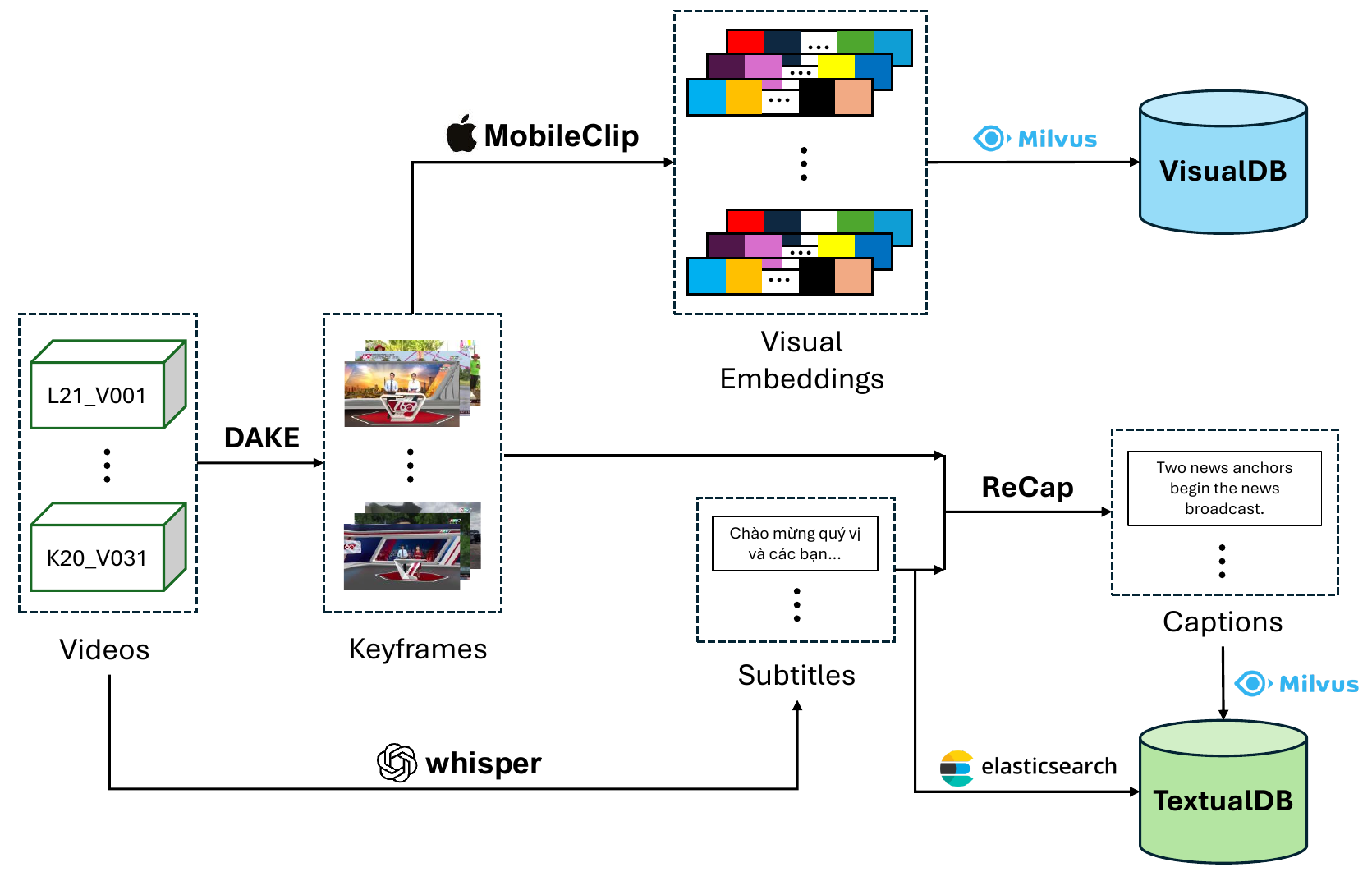}
    \caption{Our data preprocessing pipeline}
    \label{fig:data-preprocess-pipeline}
\end{figure}

\subsubsection{Dynamic-Aware Keyframe Extraction \ }

Deep learning-based methods such as AutoShot~\cite{zhuautoshot} achieve accurate shot detection but require substantial computation. Moreover, in highly dynamic scenes, AutoShot often selects only the last frame, failing to capture full motion variation. To address this, we propose \textit{Dynamic-Aware Keyframe Extraction} (DAKE), a lightweight, training-free method that leverages JPEG file sizes as motion indicators. Dynamic scenes with large motion, texture, or lighting changes exhibit abrupt variations in compressed file size~\cite{ClassicVideoShotBoundary}, while static scenes produce stable sizes. Fig.~\ref{fig:frame_size_analysis} visualizes normalized JPEG sizes across frame indices, where stable regions correspond to static scenes and sharp changes denote transitions or dynamic content. Given two frames $f_i$ and $f_j$ with corresponding JPEG file sizes $s_i$ and $s_j$, we define the \textit{steepness} $\mathcal{S}(i,j)$ as the normalized rate of change in frame size, computed as: 
\begin{equation*} \mathcal{S}(i,j) = \frac{100\times\left|\frac{s_j - s_i}{s_{\max}}\right|} {\sqrt{(j - i)^2 + \left(100\times\left|\frac{s_j - s_i}{s_{\max}}\right|\right)^2}}, 
\end{equation*} 
where $s_{\max}$ is the maximum JPEG file size in the video. This normalization ensures that steepness values remain scale-invariant across videos with varying resolutions and compression levels. To detect visual transitions, we aggregate local steepness scores within short temporal windows and select frames with the highest aggregated values, as described in Algorithm~\ref{algo:DAKE}. This approach identifies frames exhibiting strong temporal variations in compression complexity, effectively capturing shot boundaries and dynamic scenes without any learned model. Moreover, DAKE allows explicit control over the number of detected keyframes, facilitating experiments on the trade-off between keyframe recall and storage efficiency. A detailed analysis of DAKE is provided in Sec.~\ref{sec:DAKE_abl}.

\begin{algorithm}[h]
\caption{Dynamic-aware Keyframe Selection}
\label{algo:DAKE}
\begin{algorithmic}[1]
\Require List of frame sizes $\{s_1, s_2, \dots, s_n\}$, keyframe ratio $\rho$
\Ensure List of selected keyframe indices $\mathcal{K}$
\State $s_{\max} \gets \max(s_1, s_2, \dots, s_n)$
\State $\mathcal{S}_{\text{list}} \gets [~]$
\For{$i = 1$ to $n-1$}
    \State $sum \gets 0$
    \State $count \gets 0$
    \For{$j = i+1$ to $\min(n, i+3)$}
        \State $sum \gets sum + \mathcal{S}(i,j)$
        \State $count \gets count + 1$
    \EndFor
    \State $steepness \gets sum / count$
    \State Append $(i, steepness)$ to $\mathcal{S}_{\text{list}}$
\EndFor
\State Sort $\mathcal{S}_{\text{list}}$ in descending order by steepness value
\State $k \gets \lfloor \rho \times |\mathcal{S}_{\text{list}}| \rfloor$
\State $\mathcal{K} \gets$ indices of top-$k$ entries in $\mathcal{S}_{\text{list}}$
\State \Return $\mathcal{K}$
\end{algorithmic}
\end{algorithm}

\subsubsection{Keyframe Captioning \ }
\label{sec:ReCap}

We aim to generate detailed and context-aware captions for individual keyframes using a Large Vision–Language Model (LVLM) that incorporates surrounding visual and textual context. For each target keyframe $I_t$, the model receives three inputs: (1) a context window of $k$ preceding and succeeding keyframes $\{I_{t-k}, \ldots, I_{t-1}, I_t, I_{t+1}, \ldots, I_{t+k}\}$; (2) the subtitle transcript for the same segment; and (3) the target keyframe itself. The context frames and subtitle provide auxiliary cues to interpret $I_t$ but are not directly referenced in the final caption. Guided by a unified instruction prompt adapted from CESE's multi-prompt framework \cite{cese}, the LVLM integrates reasoning and caption generation within a single prompt. The instruction directs the model to:
\begin{enumerate}
    \item Infer contextual background from the context window and subtitle.
    \item Analyze visual content from entities, attributes, and interactions in $I_t$.
    \item Conduct internal reasoning via self-generated visual question–answer pairs to ensure descriptive completeness.
\end{enumerate}

The model outputs a detailed paragraph $C_t$ summarizing all important visual and contextual cues:
$$
C_t = \text{LVLM}(\{I_{t-k:t+k}\}, S_{t-k:t+k}, I_t),
$$
where $S_{t-k:t+k}$ denotes the subtitle segment. This formulation yields fine-grained, contextually grounded captions focused on the visual specificity of each keyframe. In our implementation, we employ Gemini \cite{gemini} as the underlying LVLM.

\subsubsection{Recurrence Captioning \ }

\begin{figure}[t!]
    \centering
    \includegraphics[width=1.0\linewidth]{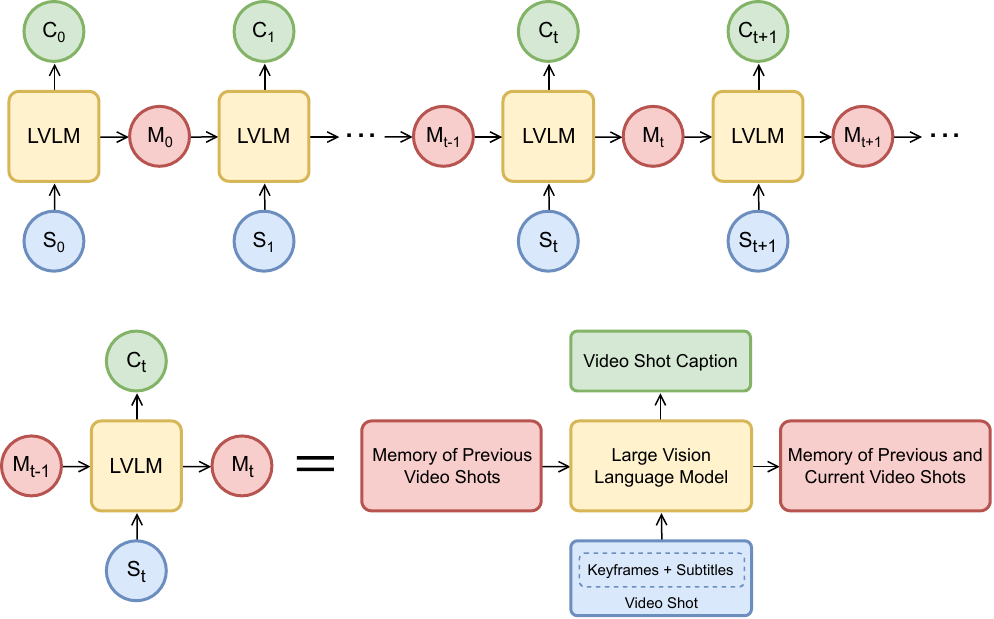}
    \caption{Overview of the Recurrence Captioning (ReCap) framework.}
    \label{fig:shot_caption_rnn}
\end{figure}

ReCap (Fig.~\ref{fig:shot_caption_rnn}) extends keyframe captioning to shot-level captioning with temporal consistency. A video is divided into shots:
$$
S = \{S_0, S_1, \ldots, S_T\},
$$
where each shot $S_t$ includes keyframes and a subtitle. We utilize AutoShot~\cite{zhuautoshot} for this task. At time step $t$, the system maintains a memory string $M_t$ capturing accumulated contextual information. 
We employ Gemini \cite{gemini} as the reasoning and generation engine:
\[
(C_t, M_t) = f_{\text{LVLM}}(S_t, M_{t-1}),
\]
where $f_{\text{LVLM}}$ denotes the LVLM’s reasoning and generation function. This recurrent design mirrors an RNN’s hidden state, preserving contextual continuity and reflecting evolving narratives across shots. The LVLM selectively retains relevant details (e.g., setting, characters, actions) while discarding outdated ones, enabling coherent captioning across scene transitions. By combining reasoning capabilities with recurrent memory, ReCap achieves temporally consistent and semantically rich video shot descriptions, bridging static image captioning and dynamic video understanding. Detailed analysis can be found in Sec~\ref{sec:ReCap_abl}.

\subsubsection{Overall Data Preprocess Pipeline \ }

Fig.~\ref{fig:data-preprocess-pipeline} illustrates our data preprocessing pipeline. Following~\cite{cese}, we first extract keyframes using DAKE. Visual embeddings are generated by MobileCLIP’s vision encoder~\cite{MobileClip} and indexed into the VisualDB by Milvus~\cite{2021milvus}. OpenAI’s Whisper~\cite{Whisper} transcribes audio into subtitles, which are combined with keyframes to create detailed captions using ReCap. Text data are indexed in raw form with Elasticsearch~\cite{elasticsearch} and as embeddings with MobileCLIP’s textual encoder in Milvus, jointly forming the TextualDB.

%% file: SystemOverview/UnifiedClippingAlgorithm/UnifiedClippingAlgorithm.tex
\subsection{Unified Clipping Algorithm}
\label{sec:unified_clipping}

Similar to CESE~\cite{cese}, we retrieve relevant clips instead of individual frames, referring to the results as "suggestions." Unlike CESE, which separates modality-specific components with distinct interfaces and ranking schemes, our \textbf{Unified Clipping Algorithm} integrates all modalities into a single retrieval and ranking pipeline, eliminating interface switching for a consistent user experience. The algorithm jointly retrieves timestamps from VisualDB (frame embeddings) and TextualDB (caption embeddings and raw text), and merges them into a single list. Given a maximum clip length $T$, a linear-time two-pointer sweep forms clips satisfying $\text{end} - \text{start} \le T$. With fixed $T$, suggestions generation and ranking times only scale linearly with the number of retrieved items $N$. Suggestions are ranked by the number of unique query IDs covered, with ties broken by the highest per-frame similarity. The full procedure is shown in Algorithm~\ref{alg:unified-clipping}.

\begin{algorithm}
\caption{Unified Clipping Algorithm}
\label{alg:unified-clipping}
\begin{algorithmic}[1]
\Require Queries $\mathcal{Q} = \{q_1,\ldots,q_n\}$; 

$\textsc{UseFrameEmb}[i], \textsc{UseTextEmb}[i], \textsc{UseTextRaw}[i]$ for each $q_i$; 

retrieval budgets $k_{\text{vis}}, k_{\text{text\_emb}}, k_{\text{text\_raw}}$; 

maximum suggestion length $T$; number of suggestions to return $K$
\Ensure Top-$K$ \texttt{suggestion}s across all videos
\Statex
\Function{RetrieveAll}{$\mathcal{Q}$}
    \State $R_{\text{vis}} \gets \texttt{VisualDB}.\textsc{retrieve\_embedding}(\mathcal{Q}, k_{\text{vis}}, \textsc{UseFrameEmb})$
    \State $R_{\text{text\_emb}} \gets \texttt{TextualDB}.\textsc{retrieve\_embedding}(\mathcal{Q}, k_{\text{text\_emb}}, \textsc{UseTextEmb})$
    \State $R_{\text{text\_raw}} \gets \texttt{TextualDB}.\textsc{retrieve\_raw}(\mathcal{Q}, k_{\text{text\_raw}}, \textsc{UseTextRaw})$
    \State $R \gets \textsc{Flatten}(R_{\text{vis}} \cup R_{\text{text\_emb}} \cup R_{\text{text\_raw}})$ \Comment{$R$ is a list of \textsc{RetrievedFrame}}
    \State \Return $\textsc{SortBy}(R,\ \langle\text{video\_name},\ \text{timestamp}\rangle)$
\EndFunction
\Statex
\Function{Better}{$S_a, S_b$} \Comment{two \texttt{suggestion}s}
    \State $U_a \gets \big|\{\text{query\_id}(f)\,:\, f \in S_a.\text{retrieved\_frames}\}\big|$
    \State $U_b \gets \big|\{\text{query\_id}(f)\,:\, f \in S_b.\text{retrieved\_frames}\}\big|$
    \If{$U_a \ne U_b$} \Return $U_a > U_b$ \EndIf
    \State $m_a \gets \max_{f \in S_a.\text{retrieved\_frames}} \text{score}(f)$
    \State $m_b \gets \max_{f \in S_b.\text{retrieved\_frames}} \text{score}(f)$
    \State \Return $m_a > m_b$
\EndFunction
\Statex
\Function{UnifiedClipping}{$\mathcal{Q}, T, K$}
    \State $R \gets \textsc{RetrieveAll}(\mathcal{Q})$
    \State $\mathcal{G} \gets \textsc{GroupBy}(R,\ \text{video\_name})$ \Comment{per-video sorted lists}
    \State $\mathcal{S} \gets [\ ]$ \Comment{candidate \texttt{suggestion}s}
    \ForAll{$\langle v, L\rangle \in \mathcal{G}$} \Comment{$L$ is sorted by timestamp}
        \State $l \gets 1$
        \For{$r \gets 1$ \textbf{to} $|L|$}
            \While{$l \le r$ \textbf{and} $L[r].\text{timestamp} - L[l].\text{timestamp} > T$}
                \State $l \gets l+1$
            \EndWhile
            \State $S \gets \textsc{Suggestion}(\text{video\_name}=v,$
            \Statex\hspace{6.1em}$\text{start}=L[l].\text{timestamp},\ \text{end}=L[r].\text{timestamp},$
            \Statex\hspace{6.1em}$\text{retrieved\_frames}=L[l{:}r])$
            \State $\mathcal{S}.\textsc{append}(S)$
        \EndFor
    \EndFor
    \State $\mathcal{S} \gets \textsc{Sort}(\mathcal{S},\ \textsc{Better})$
    \State \Return $\mathcal{S}[1{:}K]$
\EndFunction
\end{algorithmic}
\end{algorithm}

%% file: SystemOverview/UserInterface/UserInterface.tex
\subsection{User Interface}
\label{sec:user-interface}

\begin{figure}[h]
    \centering
    \includegraphics[width=0.8\linewidth]{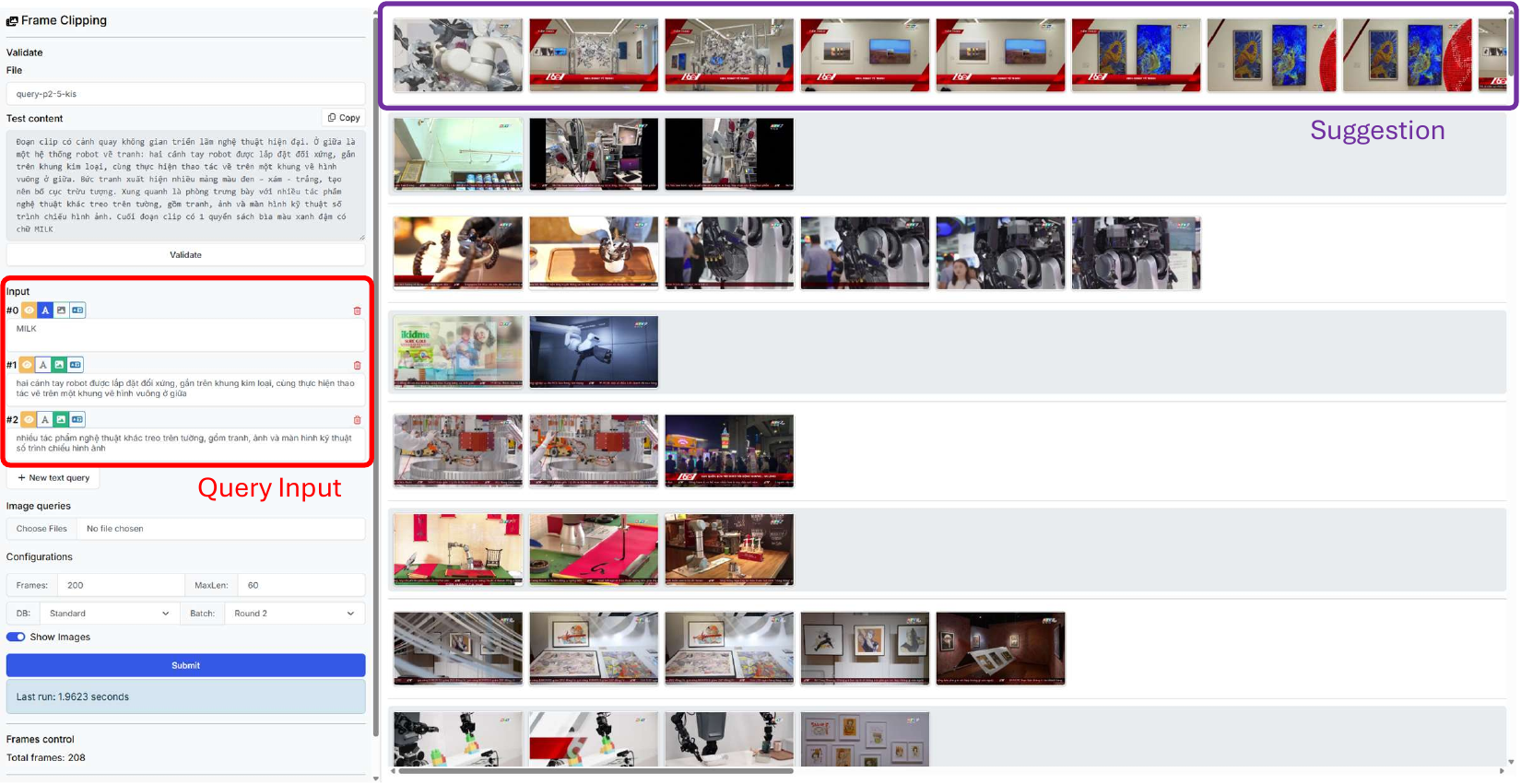}
    \caption{Main screen of U-CESE's user interface}
    \label{fig:UI-main}
\end{figure}

\begin{figure}[h]
    \centering
    \includegraphics[width=0.8\linewidth]{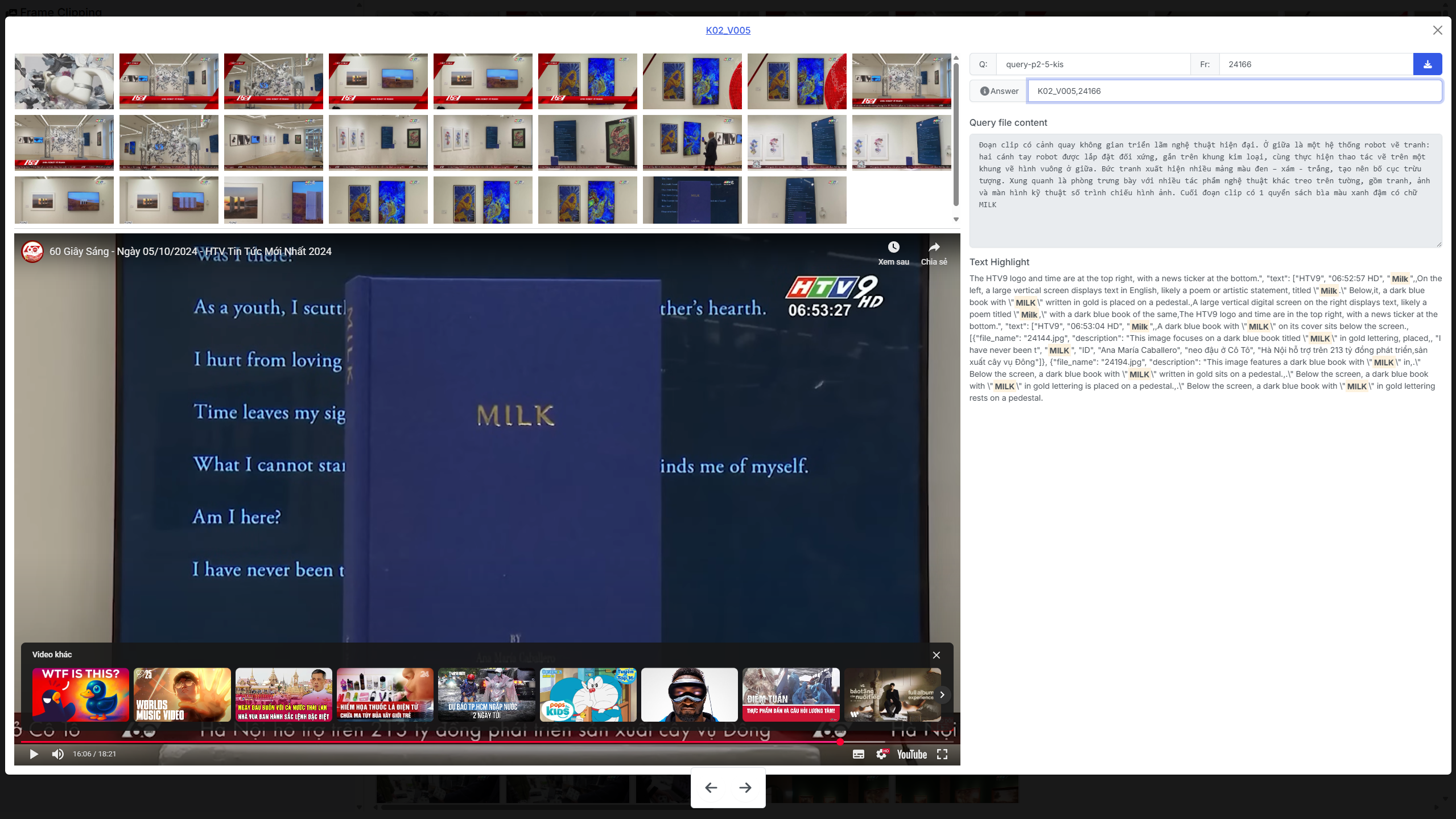}
    \caption{U-CESE's Interactive Window}
    \label{fig:UI-window}
\end{figure}

We developed a unified user interface that integrates Python, HTML, CSS, and JavaScript. As shown in Fig.~\ref{fig:UI-main}, users can input text queries or upload images for visual search. The interface allows users to view the current question, toggle active modalities, and submit their queries seamlessly. Upon submission, the system retrieves and displays a ranked list of suggestions, with each row representing one retrieved result and its associated frames. By selecting a frame, users can open an interactive viewer (Fig.~\ref{fig:UI-window}) that presents detailed information about the retrieved frames and corresponding texts. This viewer includes a built-in video player showing the current frame index in the top-right corner—an especially useful feature for TRAKE tasks that involve identifying multiple key moments in a video. Additionally, users can edit their answers in the \texttt{Answer} field and export the results as a CSV file using the download button. We also include several keyboard shortcuts to speed up common actions. For example, pressing Tab automatically appends the current timestamp to the active answer field. This is especially useful for TRAKE tasks, where users need to quickly select the exact frame corresponding to multiple events within a single query.

\vspace{-3mm}

\newpage

%% file: AblationStudy/AblationStudy.tex
\section{Ablation Study}

\subsection{Comparing DAKE with AutoShot}
\label{sec:DAKE_abl}

\begin{figure}
    \centering
    \begin{minipage}{0.58\linewidth}
        \centering
        \includegraphics[width=\linewidth]{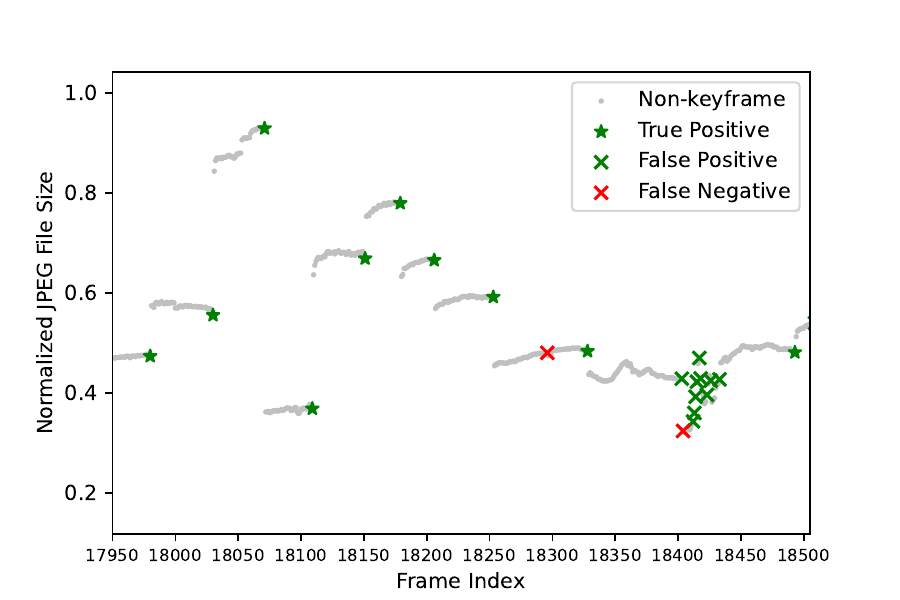}
        \vspace{-1em}
        \caption{Frame JPEG sizes across frame indices in a video sample. “True Positive” denotes exact matches, “False Positive” refers to DAKE detections not found in AutoShot results, and “False Negative” indicates keyframes detected by AutoShot but missed by DAKE.}
        \label{fig:frame_size_analysis}
    \end{minipage}\hfill
    \begin{minipage}{0.4\linewidth}
        \centering
        \includegraphics[width=\linewidth]{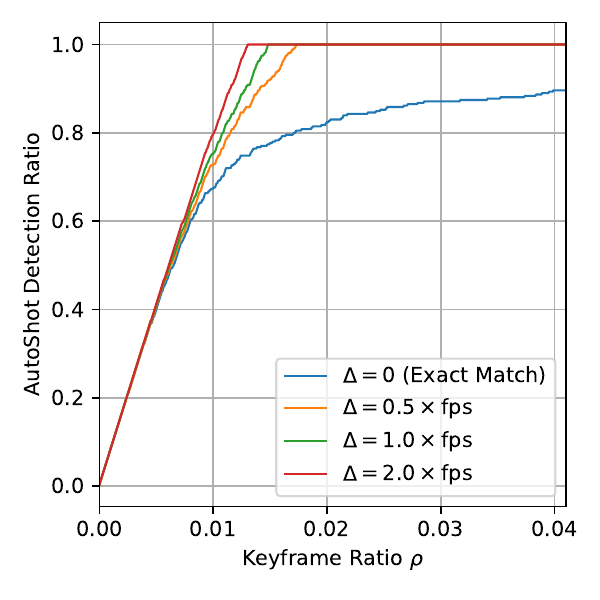}
        \caption{Coverage ratio of AutoShot detections under varying keyframe ratios $\rho$ and error windows $\Delta$.}
        \label{fig:dake_ratio}
    \end{minipage}
\end{figure}

We compare keyframes detected by DAKE with those identified by AutoShot on the video \textit{K01\_V001} from the organizers’ dataset. As shown in Fig.~\ref{fig:dake_ratio}, setting $\rho = 0.02$ (i.e., selecting only $2\%$ of all frames) captures over $80\%$ of AutoShot’s detections. Let $fps$ denote the frame rate of the video, and define a match between a DAKE detection $\hat{x}$ and an AutoShot detection $\tilde{x}$ if $\hat{x} - \Delta \leq \tilde{x} \leq \hat{x}$. With $\rho = 0.02$ and $\Delta = 0.5 \times fps$ (half a second), every AutoShot detection has at least one corresponding DAKE detection. Therefore, we adopt $\rho = 0.02$ and additionally enforce that at least one keyframe is included within every $\Delta = 2 \times fps$-frame window for a near-perfect recall.

Fig.~\ref{fig:frame_size_analysis} illustrates all methods’ detections for a representative clip. DAKE consistently covers nearly all AutoShot detections and, between frames 18400 and 18450, captures additional keyframes in highly dynamic regions. This behavior results in higher recall and mitigates the risk of losing important visual information. Since both recall and storage depend on $\rho$, the trade-off between them can be efficiently adjusted by tuning this hyperparameter. These results demonstrate DAKE’s potential as a robust and efficient shot boundary detection method, particularly suited for applications where storage and processing efficiency are critical.

\newpage

\subsection{Effect of ReCap on Captions}
\label{sec:ReCap_abl}

\begin{figure}
    \centering
    \includegraphics[width=1.0\linewidth]{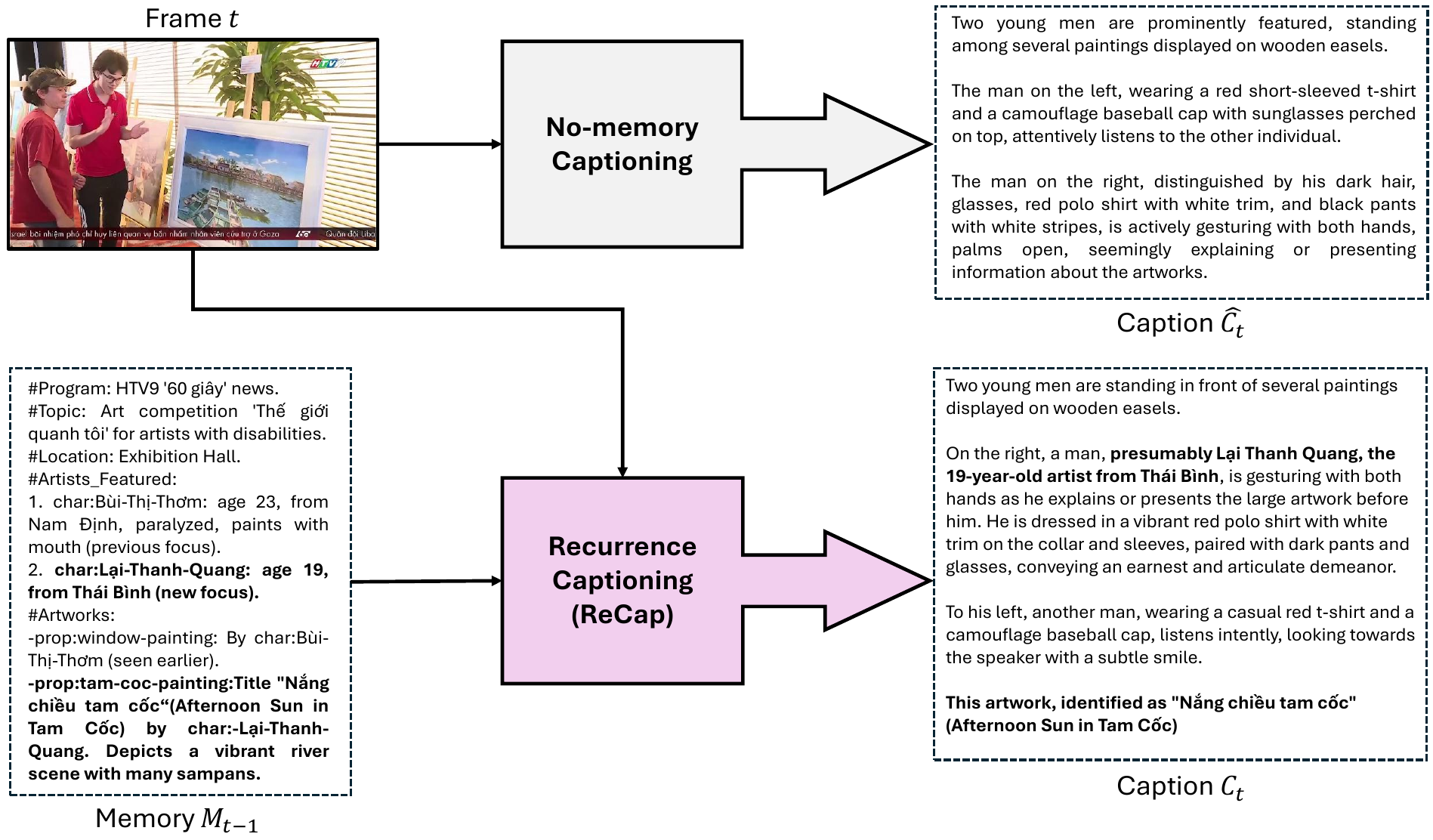}
    \caption{Effect of Recurrent Memory on captions.}
    \label{fig:ReCap_abl}
\end{figure}

We compare two configurations: (i) \emph{No-memory Captioning}, which generates captions for each shot independently using only keyframes and subtitles, and (ii) our \emph{Recurrence Captioning (ReCap)}, which conditions on the recurrent memory $M_{t-1}$ accumulated from preceding shots. All prompts, keyframe selections, and subtitle inputs remain identical across configurations; the only difference is the inclusion of $M_{t-1}$. As illustrated in Fig.~\ref{fig:ReCap_abl}, incorporating recurrent memory enables the model to maintain temporal consistency and capture richer contextual dependencies across shots.

Without memory, the model outputs $\hat{C_t}$, focusing primarily on surface-level attributes such as clothing and gestures, thereby describing individuals as generic participants. In contrast, ReCap generates $C_t$ by leveraging historical context stored in $M_{t-1}$ to construct a coherent and temporally grounded narrative. It correctly identifies the right-hand speaker as the contestant presenting his painting, provides relevant personal details mentioned in previous shots, and links the artwork to its title.

These observations demonstrate ReCap’s effectiveness in enhancing both identity tracking and contextual coherence. The recurrent memory preserves entity-specific information, preventing the model from reverting to ambiguous or generic references. Furthermore, it enables recovery of background context (e.g., program name, location, or event type) that may not be visually evident in the current frame, thereby reducing ambiguity and producing more grounded and informative captions.

\subsection{Impact of the Unified Clipping Algorithm for Event Retrieval}

In this section, we demonstrate the impact of the Unified Clipping Algorithm on U-CESE’s retrieval pipeline. Specifically, users can submit multiple queries simultaneously, each corresponding to a different data modality. As illustrated in Fig.~\ref{fig:UI-main}, users can choose between the frame modality for visual descriptive queries and the text modality for keyword-based search (e.g., \texttt{query-p2-5-kis}). The selection is performed through the frame and text icons highlighted in red. The system then generates suggestions based on multimodal retrieval items, including frames, captions, and subtitles, which are aggregated into a unified interface. When users open the a suggestion's interactive window, as shown in Fig.~\ref{fig:UI-window}, all related retrieval results are displayed cohesively, allowing seamless exploration of semantic relationships across modalities. This unified visualization encourages cross-modal comprehension, and enables users to identify contextually relevant events spanning visual and textual domains. Consequently, the Unified Clipping Algorithm not only enhances retrieval accuracy but also improves user experience by delivering an integrated and interpretable multimodal representation.

\subsection{Performance in the final round of the challenge}
\begin{figure}
    \centering
    \includegraphics[width=0.8\linewidth]{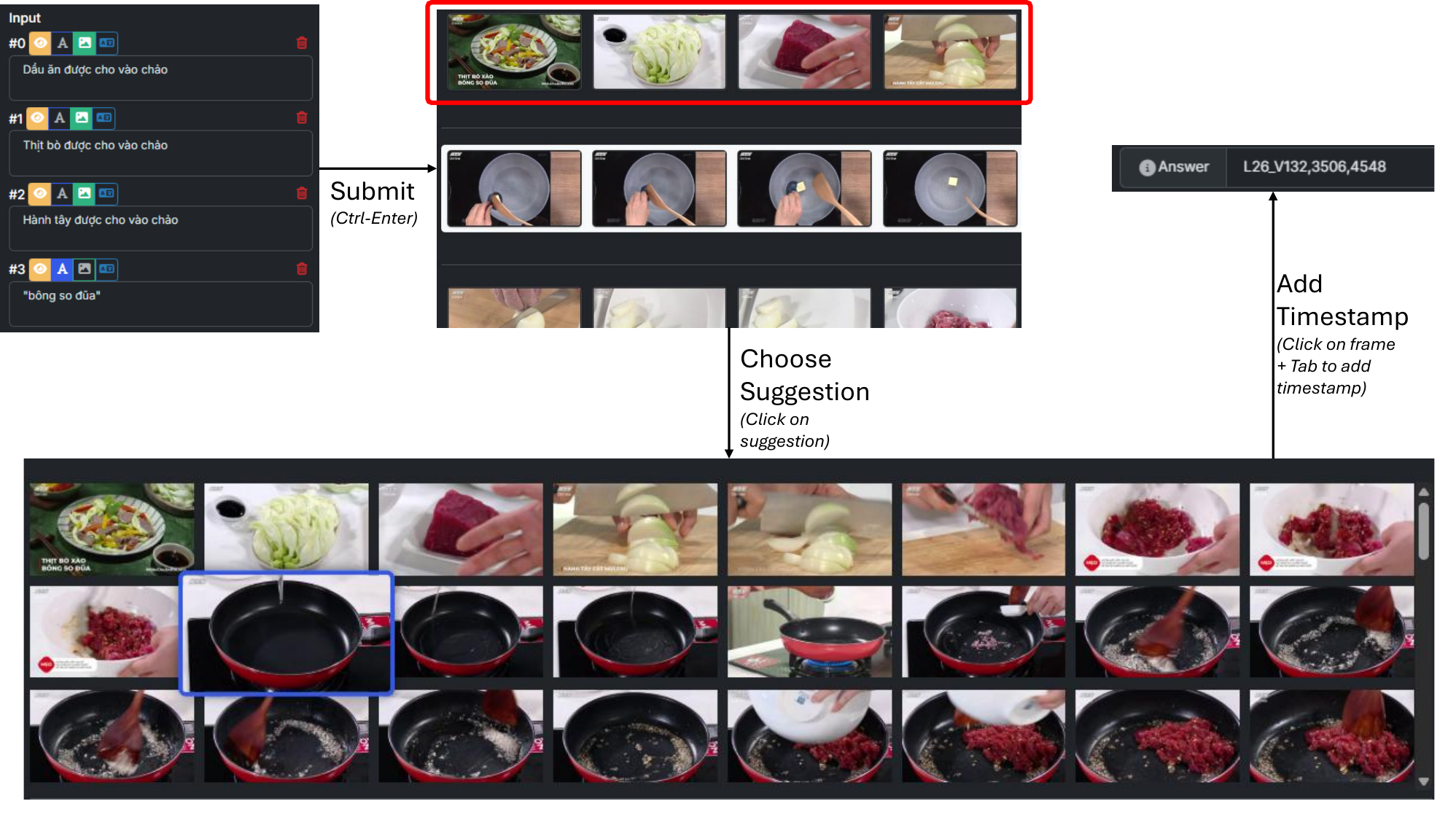}
    \caption{User's workflow for TRAKE queries. The chosen query is the 4-th TRAKE in the final round of AIC2025, which is "In a stir-fried beef cooking video, identify the first moments when each of the following ingredients makes contact with the pan: E1: cooking oil, E2: beef, E3: onion, E4: sesbania flower." By utilizing the Tab shortcut, user can quickly edit the answer.}
    \label{fig:trake-pipeline}
\end{figure}

Our team, Nomial, achieved an excellent overall performance in the final round of the AIC 2025 \cite{AIChallenge2025}. In particular, we obtained outstanding results in TRAKE and QA tasks, both of which demand a comprehensive understanding of the video dataset. In these tasks, our Unified Clipping Algorithm demonstrated its effectiveness by jointly leveraging visual and textual modalities to accurately identify the target clip. Furthermore, the inclusion of keyboard shortcuts enhances the efficiency of the TRAKE query workflow, as shown in Fig.~\ref{fig:trake-pipeline}.

%% file: Conclusion/Conclusion.tex
\section{Conclusion}
In this paper, we present \textbf{U-CESE}, a Unified Clip-based Event Search Engine developed for the AI Challenge HCMC 2025. Building upon the CESE framework, U-CESE addresses the limitations of modular inconsistency and inefficiency by integrating all modalities into a single cohesive retrieval pipeline. Our proposed \textbf{Unified Clipping Algorithm} harmonizes the clipping process across different query types, ensuring both scalability and consistency in event retrieval. To further enhance efficiency, we introduce \textbf{DAKE}, a training-free keyframe extraction method leveraging JPEG file size variance to eliminate the need for complex shot boundary models. Additionally, our \textbf{ReCap} framework generates temporally consistent and context-aware captions, enriching the semantic understanding of video content. Together, these components enable U-CESE to achieve contextually rich, effective, and scalable multimodal video retrieval across diverse, large-scale datasets. We believe that U-CESE establishes a robust foundation for future research in unified event search systems, paving the way for more advanced and intelligent video retrieval solutions.
